\documentclass[letterpaper, 10 pt, conference]{IEEEtran}
\IEEEoverridecommandlockouts                              

\usepackage{cite}
\usepackage{graphicx}
\usepackage{subcaption}
\expandafter\def\csname ver@subfig.sty\endcsname{}
\usepackage{svg}
\usepackage[version=4]{mhchem}
\usepackage{siunitx}
\usepackage{longtable,tabularx}
\usepackage{bm}
\usepackage{amsmath,amssymb,amsfonts}
\usepackage{algorithmic}
\usepackage{textcomp}
\usepackage{xcolor}
\usepackage{graphics} 
\usepackage{epstopdf} 
\usepackage{mathptmx} 
\usepackage[ruled,vlined]{algorithm2e}
\usepackage{subfig}
\usepackage{hyperref}
\usepackage{float}
\usepackage{soul}
\usepackage{breqn}
\title{A Graph-based Adversarial Imitation Learning Framework for Reliable \& Realtime Fleet Scheduling in Urban Air Mobility}

\author{Prithvi Poddar$^1$, Steve Paul$^2$ and Souma Chowdhury$^3$\\
University at Buffalo, Buffalo, NY, 14260
\thanks{$^1$Graduate Student, Department of Mechanical and Aerospace Engineering, University at Buffalo, AIAA student member. {\tt\small prithvid@buffalo.edu}}%
\thanks{$^2$Graduate Student, Department of Mechanical and Aerospace Engineering, University at Buffalo, AIAA student member. {\tt\small stevepau@buffalo.edu}}%
\thanks{$^3$Associate Professor, Department of Mechanical and Aerospace Engineering, Co-Director, Center for Embodied Autonomy and Robotics, University at Buffalo, AIAA Associate Fellow. {\tt\small soumacho@buffalo.edu}}%
}

\begin{document}
\maketitle

\begin{abstract}
The advent of Urban Air Mobility (UAM) presents the scope for a transformative shift in the domain of urban transportation. However, its widespread adoption and economic viability depends in part on the ability to optimally schedule the fleet of aircraft across vertiports in a UAM network, under uncertainties attributed to airspace congestion, changing weather conditions, and varying demands. This paper presents a comprehensive optimization formulation of the fleet scheduling problem, while also identifying the need for alternate solution approaches, since directly solving the resulting integer nonlinear programming problem is computationally prohibitive for daily fleet scheduling. Previous work has shown the effectiveness of using (graph) reinforcement learning (RL) approaches to train real-time executable policy models for fleet scheduling. {However, such policies can often be brittle on out-of-distribution scenarios or edge cases}. Moreover, training performance also deteriorates as the complexity (e.g., number of constraints) of the problem increases. To address these issues, this paper presents an imitation learning approach where the RL-based policy exploits expert demonstrations yielded by solving the exact optimization using a Genetic Algorithm. The policy model comprises Graph Neural Network (GNN) based encoders that embed the space of vertiports and aircraft, Transformer networks to encode demand, passenger fare, and transport cost profiles, and a Multi-head attention (MHA) based decoder. Expert demonstrations are used through the Generative Adversarial Imitation Learning (GAIL) algorithm. Interfaced with a UAM simulation environment involving 8 vertiports and 40 aircrafts, in terms of the daily profits earned reward, the new imitative approach achieves better mean performance and remarkable improvement in the case of unseen worst-case scenarios, compared to pure RL results.


\end{abstract}

\section{Introduction}

\begin{figure*}[h]
    \centering
    \includegraphics[width=\textwidth, trim={0.5cm 14cm 0.5cm 0.5cm},clip]{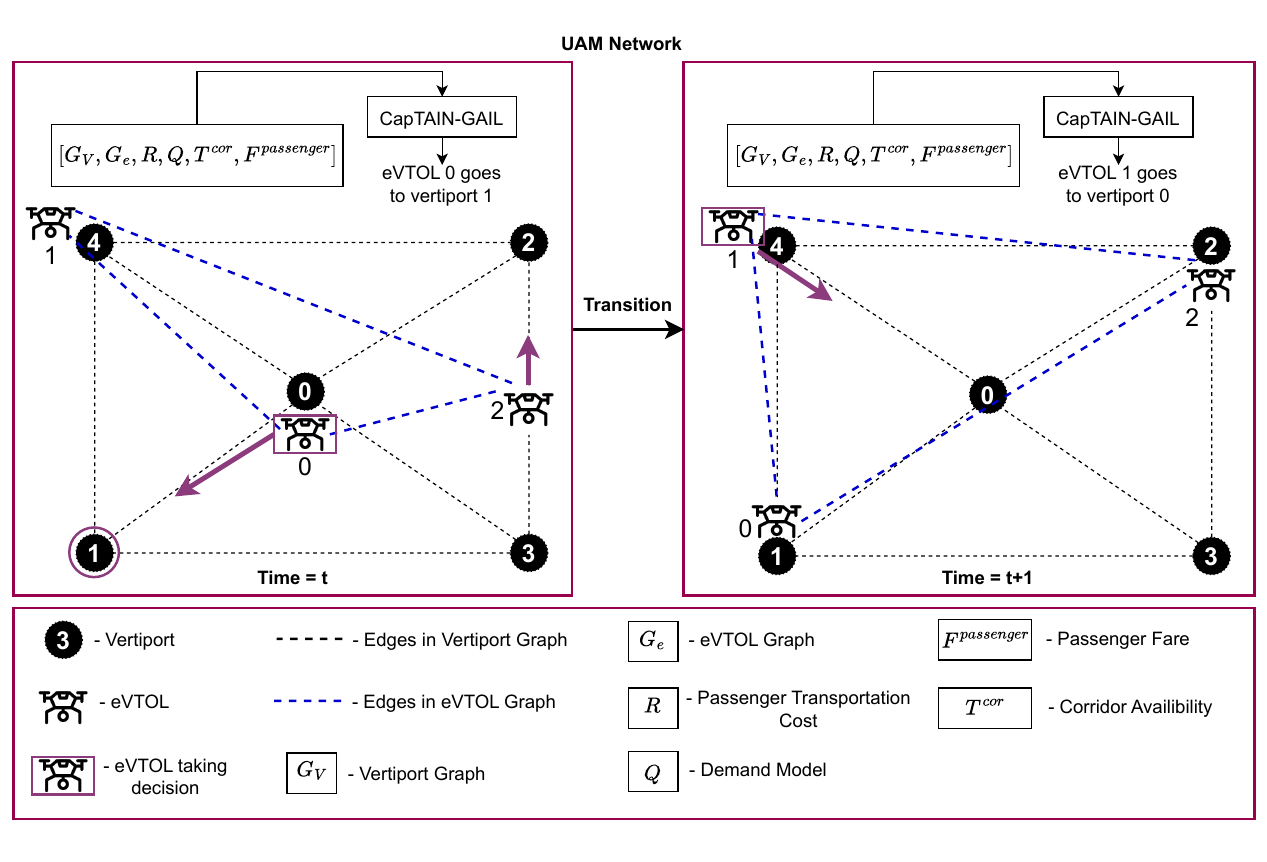}
    \caption{A visual representation of the UAM fleet management problem. CapTAIN-GAIL is the imitation learning policy that takes as input, the information from the vertiport and eVTOL graphs along with operational costs, passenger demands, and air-corridor availabilities and decides the next vertiport that the decision-making agent should go to.}
    \label{fig:inro}
\end{figure*}

In the modern landscape of urban transportation, the emergence of Urban Air Mobility (UAM) introduces a revolutionary dimension to the concept of commuting. As cities grapple with increasing populations and traffic congestion, the concept of UAM proposes the use of electric vertical take-off and landing (eVTOL) aircraft \cite{UrbanAir64:online} as an alternate form of automated air transportation. With a projected market size of $\$1.5$ trillion by 2040 \cite{INRIXGlo40:online}, its economic viability will be driven by the ability to operate a sufficiently large number of eVTOLs in any given market (high-penetration), placing the domain of UAM fleet scheduling at the forefront of innovations. Following the air-space and aircraft safety constraints while being robust against dynamic environmental changes, mitigating energy footprint, and maximizing profitability makes developing an optimal scheduling policy challenging. The complexities inherent in UAM fleet scheduling are characterized by:

\textbf{Dynamic environments} where real-time factors such as airspace congestion, weather conditions, demand uncertainty characterized by dynamic traffic patterns, changing weather conditions, and regulatory constraints demand intelligent scheduling solutions that can swiftly adapt to real-time challenges \cite{UrbanAir64:online}.

\textbf{State information sharing} among all vertiports and eVTOLs is necessary for trajectory and speed adjustments to ensure the safety of the operations \cite{Krivonos2007COMMUNICATIONIA}.

\textbf{Scarce resources} such as vertiport parking lots, charging stations, and air corridors must be optimally allocated to prevent unnecessary delays and conflicts \cite{resource_alloc}.

Contemporary solutions to such scheduling problems take the form of complex nonlinear Combinatorial Optimization (CO) problems \cite{Hwang2001}, which can be addressed through classical optimization, heuristic search, and learning-based approaches. While these approaches provide local optimal solutions for small UAM fleets \cite{8569225, 8901431, doi:10.2514/6.2020-2906}, they often present computational complexities that render them impractical for online decision-making and larger fleets. Furthermore, despite the computational expenses, these methods would still be viable options if we disregard uncertainties in the eVTOL's operations (e.g. occasional failures of the aircraft) and environmental constraints (like the closure of air corridors due to bad weather conditions). However, these factors cannot be disregarded because of the safety concerns they impose, making it especially challenging to use such approaches.

In the pursuit of addressing these complexities and presenting an efficient online scheduler, we extend our previous work in \cite{10.1145/3605098.3635976} by proposing a specialized Graph Neural Networks (GNN) \cite{10.1115/DETC2022-90123, Paul2022LearningSP} based adversarial imitation learning framework {that learns from the optimal schedules (expert data) generated by classical optimization algorithms and builds upon existing multi-agent task allocation methods for scaling up to larger UAM fleets. We begin by posing the UAM fleet scheduling problem as an Integer Non-Linear Programming (INLP) problem with a small number of eVTOLs and vertiports and generate optimal solutions using an elitist Genetic Algorithm. These form the expert demonstrations for the imitation learning policy. We use the Generative Adversarial Imitation Learning (GAIL) \cite{gail} algorithm to train the GNN policy. Then, we establish how a graph representation efficiently encodes all the state-space information of the vertiports and the eVTOLs, thus motivating the use of GNNs over regular neural networks. We also present a comparative study between a pure reinforcement learning-based method and a Genetic Algorithm, that demonstrates edge cases where standard RL methodologies fail to generate an optimal solution and require access to expert data to train an optimal policy.} Finally, we compare our results with standard RL-based and Genetic Algorithm based solutions and demonstrate that our method performs better than RL-based methods, in terms of profitability and while it performs equally to a Genetic Algorithm, it is significantly faster in it terms of computational times when compared to the Genetic Algorithm. Fig.\ref{fig:inro} presents a diagrammatic representation of the UAM fleet management problem.

\section{Related Works}

The existing body of work in Urban Air Mobility (UAM) fleet planning has witnessed notable growth, propelling the optimization and learning formalism to advance this emerging concept. However, a considerable portion of this literature neglects crucial guidelines outlined by the FAA \cite{UrbanAir64:online} concerning UAM airspace integration, particularly in aspects such as air corridors, range/battery constraints, and unforeseen events like route closures due to adverse weather or eVTOL malfunctions \cite{inproceedings,8901431,osti_10345357}. Traditional methods, including Integer Linear Programming (ILP) and metaheuristics, prove inadequate for efficiently solving related NP-hard fleet scheduling problems \cite{7294146, Miller1960IntegerPF,10.1007/3-540-55027-5_23,Peng2021, Rizzoli2007,7462285}. For context, we focus on hour-ahead planning to allow flexibility in adapting to fluctuating demand and route conditions affected by weather and unforeseen aircraft downtime. While learning-based methods have shown promise in generating policies for combinatorial optimization problems with relatable characteristics \cite{Barrett_Clements_Foerster_Lvovsky_2020, Khalil2017LearningCO,9750805, article,Kool2018AttentionLT,NEURIPS2018_8d3bba74,nowak2018revised,Paul2022LearningSP}, their current forms often oversimplify complexities or tackle less intricate problem scenarios. In response, our paper introduces a centralized learning-based approach tailored to address the complexities and safety guidelines inherent in UAM fleet scheduling. Our proposed approach involves the creation of a simulation environment that encapsulates these complexities, facilitating the training of a policy network for generating hour-ahead sequential actions for eVTOLs across a generic 12-hour operational day. The policy network integrates a Graph Capsule Convolutional Neural Network (GCACPN) \cite{gcaps,JacobPaulNature2024} for encoding vertiport and eVTOL state information presented as graphs, a Transformer encoder network for assimilating time-series data on demand and fare, and a feedforward network for encoding passenger transportation cost. Additionally, a Multi-head Attention mechanism is employed to fuse the encoded information and problem-specific context, enhancing sequential decision-making \cite{Kool2018AttentionLT,NIPS2017_3f5ee243}.

\section{Problem Description and Formulation}

\subsection{UAM Fleet Scheduling as an Integer Non-Linear Programming Problem} \label{sec:INLP_formulation}

We begin by posing the UAM fleet scheduling problem as an optimization problem \cite{10.1145/3605098.3635976}, enabling us to use a Genetic Algorithm to generate the expert solution. Consider a UAM network comprising $N$ vertiports and $N_K$ number of eVTOLs, each with a maximum passenger carrying capacity of $C(=4)$. Let $V$ and $K$ denote the set of all vertiports and eVTOLs, respectively, with each vertiport $i\in V$ capable of accommodating a maximum of $C_{max}^{park}(=10)$ number of eVTOLs while also having charging stations for $C_{max}^{charge}(=6)$ eVTOLS. Some vertiports might not have charging stations and are called vertistops $(V_S\subset V)$. We consider there to be 4 air corridors between two vertiports with two corridors for each direction. Each vertiport $i \in V$ has an expected take-off delay $T_i^{TOD}$ that affects every eVTOL taking off, and we consider this estimated take-off delay to be less than 6 minutes at each vertiport. The probability of route closure between any two vertiports $i$ and $j$ is considered to be $P_{i,j}^{closure}(\leq 0.05)$ while the probability of an eVTOL $k$ becoming dysfunctional is considered to be $P_k^{fail}(\leq 0.005)$. Every episode considers a randomly assigned take-off delay ($\leq 30$ mins) that is drawn from a Gaussian distribution with mean $T_i^{TOD}$ and a standard deviation of 6 minutes. Additional assumptions that are made regarding the problem setup are: 1) The decision-making is done by a central agent that has access to the full observation of the states of the eVTOLs and the vertiports, and 2) An eVTOL can travel between any two vertiports given it has sufficient battery charge and the resistive loss of batteries is negligible.

We define $R_{ij}$ as the operation cost of transporting a passenger from vertiport $i$ to $j$ while $F_{ijt}^{passenger}$ is the price a passenger is charged for traveling from $i$ to $j$ at time $t$, that is based on passenger demand. We follow the demand model as mentioned in our previous work \cite{10.1145/3605098.3635976}. Let $Q(I,j,t)$ represent the demand between vertiports $i$ and $j$ at time $t$. A subset of vertiports $V_B\subset V$ are considered to be high-demand vertiports and we account for two peak hours of operation: $8:00-9:00$ am ($T^{peak1}$) and $4:00-5:00$ pm ($T^{peak2}$) such that $V_B$ experience high demands during both peak hours. The demand is high from $V-V_B$ to $V_B$ during $T^{peak1}$ and vice-versa during $T^{peak2}$. The passenger fare is computed by adding a variable fare $F^{passenger}$ to a fixed base fare of $F^{base}=\$5$. The variable fare between two vertiports $i$ and $j$ is dependent on the passenger demand $Q(i,j,t)$ and the operational cost $R_{i,j}$ and is computed as $F_{i,j,t}^{passenger}=R_{i,j}\times Q_{factor}(i,j,t)$, where $Q_{factor}(i,j,t)=\text{max}(\log (Q_(i,j,t)/10),1)$. A constant electricity pricing, $Price^{elec}=\$0.2$/kWh is used \cite{Electric57:online}.

The eVTOL vehicle model is considered to be the City Airbus eVTOL \cite{CityAirb36:online} with a battery model similar to the one described in \cite{doi:10.2514/6.2020-2906}. The aircraft has a maximum cruising speed of 74.5 mph and a passenger seating capacity of 4. The operating cost of this vehicle is $\$0.64$ per mile \cite{doi:10.2514/6.2020-2906}. It has a maximum battery capacity of $B_{max}=110$ kWh. Let $B_t^k$ represent the charge left in eVTOL $k$ at time $t$ and if the eVTOL travels from vertiport $i$ to $j$, then the charge at the next time step is $B_{t+1}^k=B_t^k-B_{i,j}^{charge}$, where $B_{i,j}^{charge}$ is the charge required to travel from $i$ to $j$.

We consider an operating time horizon of $T$ hours (with start time $T^{start}$ and end time $T^{end}$) and the aim of the UAM scheduling problem is to maximize the profits earned in the $T$ hours. This is achieved by smartly assigning (based on the demand, battery charge, operation cost, etc.) a journey to an eVTOL during each decision-making instance $t\in T$. A journey is defined as the commute of an eVTOl between two vertiports $i$ and $j$. If $i=j$, the eVTOL has to wait for $T^{wait}(=15)$ minutes at the vertiport before assigning a journey in the next decision-making instance. We assume that each eVTOL can take off at any time within the $T$ horizon, and we consider the schedule for a single-day operation where the operational time begins at 6:00 AM ($T^{start}$) and ends at 6:00 PM ($T^{end}$).

\textbf{Optimization formulation:} Based on the notations described above, the optimization problem is formulated as follows (similar to our previous work \cite{10.1145/3605098.3635976}). For every eVTOL $k\in V_e$, let $S_k^{jour}$ be the set of journeys taken during time period of $T$ and, $N_{i,l}^{passengers}$ be the number of passengers transported during the trip, $l\in S_k^{jour}$ by eVTOL $k$. Let $B_l^k$ be the battery charge of eVTOL $k$ just before its $l$’th journey, $V_{k,l}^{start}$ and $V_{k,l}^{end}$ be the respective start and end vertiports of eVTOL $k$ during it $l$’th journey, $T_{k,l}^{takeoff}$ and $T_{k,l}^{landing}$ be the corresponding takeoff time and landing time. Then the objective of the optimization problem is to maximize the net profit $z$ that can be computed by subtracting the operational costs $C^O$ and the cost of electricity $C^E$ from the revenue generated. These values are computed as:

\begin{gather}
    C^{O} = \sum_{k\in K}\sum_{l\in S_k^{jour}} N_{i,l}^{passengers}\times R_{i,j} \hspace{0.3cm},i=V_{k,l}^{start}, j=V_{k,l}^{end}\\
    C^{E} = \sum_{k\in K}\sum_{l\in S_k^{jour}} Price^{elec}\times B_{i,j}^{charge} \hspace{0.3cm}
    ,i=V_{k,l}^{start}, j=V_{k,l}^{end} \\
    \begin{split}
        \text{Revenue} = \sum_{k\in K}\sum_{l\in S_k^{jour}} N_{i,l}^{passengers}\times F_{i,j,t}^{passenger} \hspace{0.3cm}\\
    ,i=V_{k,l}^{start}, j=V_{k,l}^{end}
    \end{split}
\end{gather}





\noindent Therefore, the objective function can be formulated as:
\begin{equation}
    \max z=\text{Revenue}-C^O-C^C
\end{equation}

With the following constraints:
\begin{gather}
    \begin{split}
        N_{i,l}^{passengers}=min(C,Q_{act}(I,j,t)) \hspace{0.3cm}\\
        ,i=V_{k,l}^{start}, j=V_{k,l}^{end}, t=T_{k,l}^{takeoff} \forall l\in S_k^{jour}
    \end{split}\\
    \begin{split}
        B_{T_{k,l}^{takeoff}}^k>B_{i,j}^{charge} \hspace{0.3cm}\\
        ,i=V_{k,l}^{start}, j=V_{k,l}^{end}, \forall k\in K, \forall l\in S_k^{jour}
    \end{split}\\
    C_{i,t}^{park}\leq C_{max}^{park}\hspace{0.3cm},\forall i\in V, \forall t \in [T^{start},T^{end}]\\
    V_{l,k}^{end}\neq i\hspace{0.3cm}, \text{if} A^{i,V_{l,k}^{end}}=0,\forall i\in V, \forall k \in K, \forall l \in S_k^{jour}
\end{gather}

Where $Q_{act}(I,j,t)$ is the actual demand, and $A$ is a matrix that represents the availability of routes between vertiports. $A_{ij}=A_{ji}=1$ if the route between $i$ and $j$ is open and $=0$ if the route is closed.



\subsection{Generating the Expert Demonstrations} \label{sec:GA}

Given the INLP problem formulation in Sec.\ref{sec:INLP_formulation}, we use an elitist Genetic Algorithm (GA) to generate the expert solutions for a small number of eVTOLs ($N_k = 40$) and vertiports ($N = 8$). The GA uses a population size of 100, max iterations of 100, mutation probability of 0.1, elite ratio of 0.01, and cross-over probability of 0.5. The GA is implemented in batches of 60 decision variables which are then simulated sequentially and the objective functions are computed. Upon simulating a batch of the GA solution, we achieve an updated state of the environment which is then used to compute the next 60 decision variables. This process is repeated until the episode is over. Each decision variable takes an integer value between 1 and $N$. We generate the GA solutions for 100 different scenarios (for each scenario, the locations of the vertiports stay fixed) which are then used as expert demonstrations for the imitation learning algorithm.

\subsection{MDP Formulation}
Having expressed the UAM fleet scheduling problem as an INLP problem, we now define it as a Markov Decision Process (MDP) that sequentially computes the action for each eVTOL at the decision-making time instant $t\in T$.  The state, action, reward, and the transitions are described below:

\textbf{State Space:} The state-space at time $t$ consists of information about \textbf{1)} the vertiports, represented as a graph $G_V$, \textbf{2)} the eVTOLs, represented as a graph $G_e$, \textbf{3)} the passenger demand model $Q$, \textbf{4)} passenger fare $F^{passenger}$, \textbf{5)} operational costs $R$ (computed based on the per mile operational cost of the eVTOL, the passenger demands and the price for electricity), and \textbf{6)} time when it is safe to launch an eVTOL to the corridors $T^{cor}\in \mathbb{R}^{N\times N\times 2}$.

 At any decision-making time step, the state information is processed by a transformer-aided multihead-attention graph capsule convolutional neural network (that was presented in  \cite{10.1145/3605098.3635976}) that acts as the policy network for the imitation learning algorithm. Details about the policy network have been discussed in \ref{sec:CAPTAIN}. Further details on the state space are as follows:

 \textit{Graph Representation of the Vertiport Network:} $G_V=(V_v, E_v, A_v)$ represents the vertiport network as a graph where $V_v (=V)$ is the set of vertiports, $E_v$ represents the set of edges/routes between the vertiports, and $A_v$ represents the adjacency matrix of the graph. To consider the route closure probability, a weighted adjacency matrix is computed as $A_v = (1_{N \times N} - P^{closure}) \times A$. The node properties $\delta^t_i$ of the vertiport $i\in V_v$ at time $t$ are defined as $\delta_i^t=[x_i,y_i,C_{i,t}^{park}, T_i^{charge}, T^{TOD}_i, I_i^{vstop}]$, where $(x_i,y_i)$ are the x-y coordinates of the vertiport, $C_{i,t}^{park}$ is the number of eVTOLs currently parked, $T_i^{charge}$ is the earliest time when a charging station will be free, $T^{TOD}_i$ is the expected take-off delay, and $I_i^{vstop}$ is a variable that takes the value 1 if the node is a vertistop and 0 if the node is a vertiport.

 \textit{Graph Representation of the eVTOLs:} $G_e = (V_e, E_e, A_e)$ represents the graph that encodes the information about the eVTOL network. $V_e$ represents the set of eVTOLs, $E_e$ represents the set of edges, and $A_e$ represents the adjacency matrix. $G_e$ is considered to be fully connected and each eVTOL $k \in V_e$ is represented by it properties i.e. $\psi_k^t = \left[ x_k^d, y_k^d, B_t^k, T_k^{\text{flight}}, T_k^{\text{dec}}, P_k^{\text{fail}} \right]$, $\psi_i^t \in \mathbb{R}^6$. Here, $x_k^d, y_k^d)$ represent the coordinates of the destination vertiport, $B_t^k$ is the current battery level, $T_k^{\text{flight}}$ is the next flight time, $T_k^{\text{dec}}$ is the next decision making time, and $P_k^{\text{fail}}$ is the probability of failure.

\textbf{Action Space:} At each decision-making time instance, each agent takes an action from the available action space. The action space consists of all the available vertiports. Therefore the action space will be of size $N_K$. During a decision-making step, if an agent chooses the vertiport at which it currently is, then it waits for 15 minutes in the vertiport, until it makes a new decision.

\textbf{Reward:} We consider a delayed reward function where the agent gets a reward only at the end of an episode. The reward is ratio of the profit earned to the maximum possible profit in an episode, i.e. $\sum_{i\in V,j\in V,t\in[T^{start},T^{end}]}(Q(i,j,t)\times F^{passenger}_{i,j,t})$.

\textbf{Transition Dynamics:} Since demand and electricity pricing can vary from that of the forecasted values, the transition of the states is considered to be stochastic. The transition is an event-based trigger. An event is defined as the condition that an eVTOL is ready for takeoff. As
environmental uncertainties and communication issues (thus partial observation) are not considered in this paper, only
deterministic state transitions are allowed.

\section{Proposed Graph-based Adversarial Imitation Learning Approach} \label{sec:methodology}

\begin{figure*}[h]
    \centering
    \includegraphics[width=\textwidth, trim={1.1cm 16cm 0 0.5cm},clip]{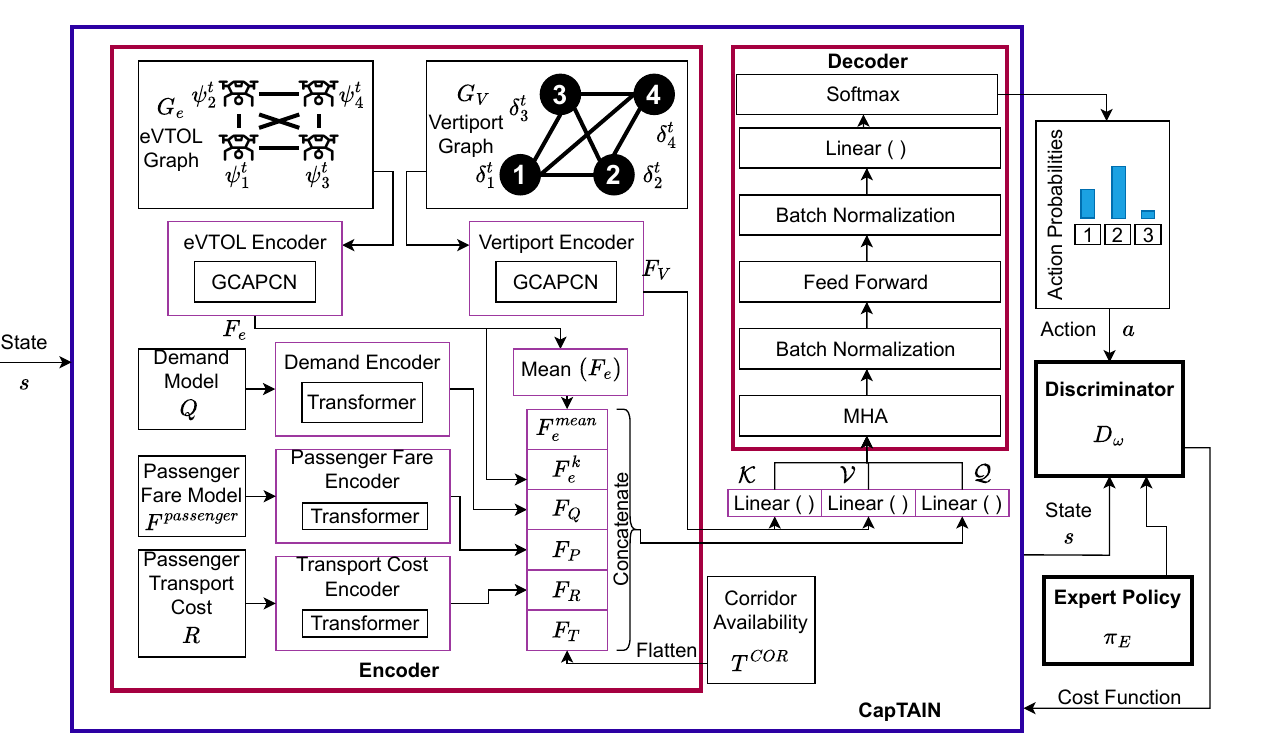}
    \caption{The learning framework for CapTAIN-GAIL}
    \label{fig:framework}
\end{figure*}

We propose an adversarial imitation learning formulation to train a transformer-aided GNN-based policy network called CapTAIN \cite{10.1145/3605098.3635976} that will assign actions to the eVTOLs. The policy network takes the state information from all the eVTOLs and vertiports as input and assigns an action to the eVTOLs that are ready for take-off. The policy is trained using the generative adversarial imitation learning (GAIL) \cite{gail} algorithm that uses the solutions generated by the GA as the expert demonstrations. The following section presents further information about the GAIL algorithm and the state encoding and decoding.

\subsection{Policy Network} \label{sec:CAPTAIN}

We use the \textbf{Cap}sule \textbf{T}ransformer \textbf{A}ttention-mechanism \textbf{I}ntegrated \textbf{N}etwork (CapTAIN) \cite{10.1145/3605098.3635976} as the policy network for GAIL. CapTAIN combines graph neural networks and transformers to encode the state-space information and used a multi-head attention-based decoder to generate the action.

The information from the veriports and eVTOLs ($G_V$ and $G_e$) are first passed through a Graph Capsule Convolutional Network (GCAPCN), introduced in \cite{gcaps}, which takes the graphs as the inputs and generates the feature embeddings for the vertiports and eVTOLs. Simultaneously, a transformer architecture is used to compute learnable feature vectors for the passenger demand model and the passenger fares (information that can be represented as time-series data). Additionally, a simple feed forward network computes the embeddings for the passenger transportation cost $R$ which can be represented as a $N\times N$ matrix, and another feed forward network processes the corridor availability. We keep a track of the time at which it is safe for a new eVTOL to enter a corridor, subject to various safety restrictions and minimum separation, using a $N\times N\times 2$ tensor,  $T^{cor}$, which is flattened and fed into the feed forward network. The outputs from the transformer, GCAPCN, and the feed forward networks form the encoded embedding which is then passed through a multi-head attention \cite{NIPS2017_3f5ee243} decoder to generate the probabilities of choosing each action, for the decision-making agent. Further technical details about CapTAIN have been discussed in \cite{10.1145/3605098.3635976}.

\subsection{Training with Generative Adversarial Imitation Learning}

The policy network $\pi$ is trained on the expert demonstrations generated by the Genetic Algorithm, using Generative Adversarial Imitation Learning (GAIL), an imitation learning approach introduced in \cite{gail}. The policy is learnt using a two-player zero-sum game which can be defined as:
\begin{equation}
\begin{split}
    \underset{\pi}{\mathrm{argmin}}\;\underset{D\in(0,1)}{\mathrm{argmax}} \; \mathbb{E}_\pi\left[\log{D(s,a)}\right]+
    \\\mathbb{E}_{\pi_E}\left[\log{(1-D(s,a))}\right]-\lambda H(\pi)
\end{split}
\end{equation}

\noindent Where $\pi_E$ is the policy followed by the expert demonstration. $D$ is a discriminator that solves a binary classification problem $D:\mathcal{S}\times \mathcal{A}\rightarrow (0,1)$, where $\mathcal{S}$ represents the state-space and $\mathcal{A}$ represents the action-space. $D$ tries to distinguish between the distribution of data generated by $\pi$ from the expert data generated by $\pi_E$, When D cannot distinguish data generated by the policy $\pi$ from the true data, then $\pi$ has successfully matched the true data. $H(\lambda)=\mathbb{E}_{\pi}\left[ -\log{\pi(a|s)}\right]$ is the entropy. Figure \ref{fig:framework} shows the overall learning framework.

The discriminator (approximated by a neural network) tries to maximize the objective, while the policy $\pi$ tries to minimize it. GAIL alternates between an Adam gradient [6] step on the parameters $w$ of $D_w$ with the gradient:
\begin{equation}
\hat{\mathbb{E}}_{\tau_i}\left[\nabla_w\log{D_w(s,a)}\right]+\hat{\mathbb{E}}_{\mathcal{D}}\left[\nabla_w\log{1-D_w(s,a)}\right]
\end{equation}
and a Trust Region Policy Optimization (TRPO) [7] gradient step on the parameters $\theta$ of $\pi_theta$ which minimizes a cost function $c(s, a) = \log D_{w_{i+1}}(s, a)$ with the gradient:

\begin{equation}
    \hat{\mathbb{E}}_{\tau_i}\left[\nabla_\theta\log{\pi_\theta}(a|s)Q(s,a)\right]-\lambda\nabla_\theta H(\pi_\theta)
\end{equation}

\section{Experiments and Results}

\subsection{Simulation Details}
We use the simulation environment presented in \cite{10.1145/3605098.3635976} that is implemented in Python and uses the Open AI Gym environment interface. We consider a hypothetical city covering an area of $50\times50$ sq. miles with 8 vertiports (2 of which are vertistops) and 40 eVTOLs. The locations of the vertiports remain the same throughout training and testing the algorithm. The rest of the operational details stay the same, as discussed above. To train the CapTAIN policy network, we use the GAIL implementation from \textbf{imitation} \cite{gleave2022imitation}, a python package for imitation learning algorithms, and PPO \cite{Schulman2017ProximalPO} from \textbf{stable-baselines3} \cite{stable-baselines3}.

\subsection{Training Details}

We begin by training the \textbf{CapTAIN} policy using just PPO, for 1.5 million steps. Next, we generate the expert demonstrations for 100 scenarios (un-seen by CapTAIN during training) using a standard elitist Genetic Algorithm (\textbf{GA}) (using the parameters as described in {Sec.\ref{sec:GA}}) and compare the performance of CapTAIN against GA. These test scenarios are generated by fixing the seed values for the random number generators in the simulation. Finally, we train the imitation learning policy (\textbf{CapTAIN-GAIL}) on the expert demonstration, with a soft start by beginning the training with the pre-trained weights from CapTAIN, and finally compare the performance of CapTAIN-GAIL against CapTAIN and GA. All the training, experiments, and evaluations are computed on a workstation running Ubuntu 22.04.4 LTS and equipped with an Intel Core i9-12900K processor and Nvidia GeForce RTX 3080 Ti graphics processor. Further evaluation details and results are discussed in the sections ahead.

\begin{figure}[hbt!]
    \centering
    \includegraphics[width=8.5cm]{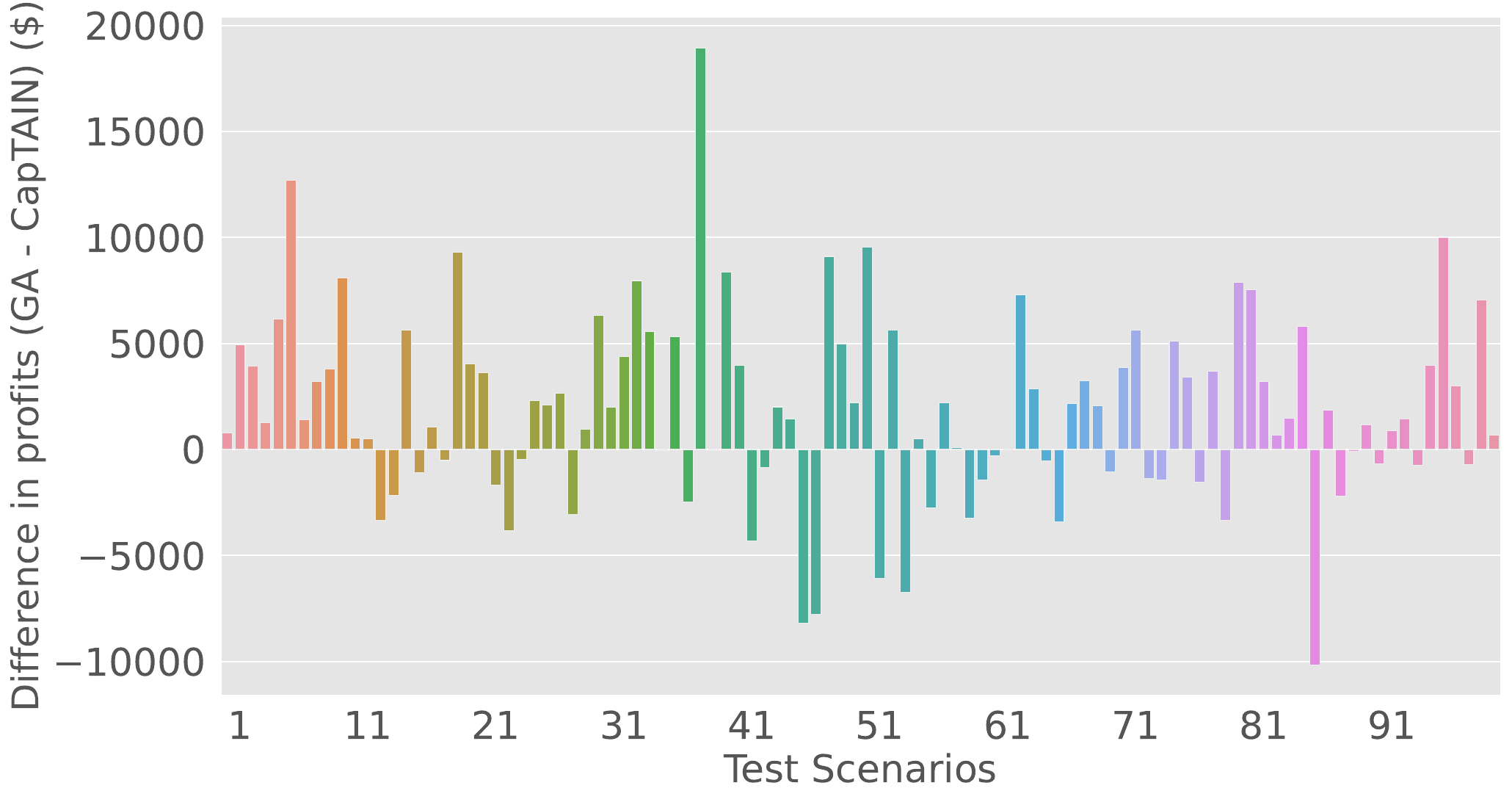}
    \caption{Difference in the profits earned by GA and CapTAIN, Positive values indicate better performance by GA and negative values indicate better performance by CapTAIN.}
    \label{fig:rl_vs_ga}
\end{figure}

\subsection{Comparing CapTAIN vs. Genetic Algorithm} \label{sec:rl_vs_ga}
We begin by testing the performance of a policy trained purely using reinforcement learning (i.e. CapTAIN trained using PPO) against GA. The daily profit earned by both the algorithms in each test scenario, is used to compare their performance and Fig.\ref{fig:rl_vs_ga} plots the difference in the profits earned by GA and CapTAIN in each of the 100 test scenarios.

It can be seen that GA generates more profits in the majority of scenarios (\textbf{GA performs better in 66 out of 100 cases}), with \textbf{GA generating an average profit of $\bm{\$17603}$ while CapTAIN generating an average profit of $\bm{\$15727}$.} To test the significance of this difference, we perform a statistical T-test with the null-hypothesis being that both methods generate the same amount of profit. The \textbf{p-value of the test turns out to be} $\bm{3.08\times 10^{-5}(<0.05)}$ which means that GA has a significant statistical advantage over CapTAIN. This motivates the use of imitation learning for training a policy that can mimic GA while being significantly more efficient than GA in terms of computational time.

\subsection{Evaluating CapTAIN-GAIL}
We train the imitation learning policy with a soft start, i.e. we use the pre-trained CapTAIN policy from Sec.\ref{sec:rl_vs_ga} as the initial policy for GAIL. The generator policy in GAIL is trained for 20 iterations with 20000 steps in each iteration. The trained imitation learning policy is then tested on the 100 test cases and its performance is compared against CapTAIN and GA. We further test the generalizability of CapTAIN-GAIL on a new set of 100 unseen test cases that were not a part of the expert demonstrations.

\subsubsection{Average Profits}

\begin{figure}[hbt!]
    \centering
    \includegraphics[width=8.5cm]{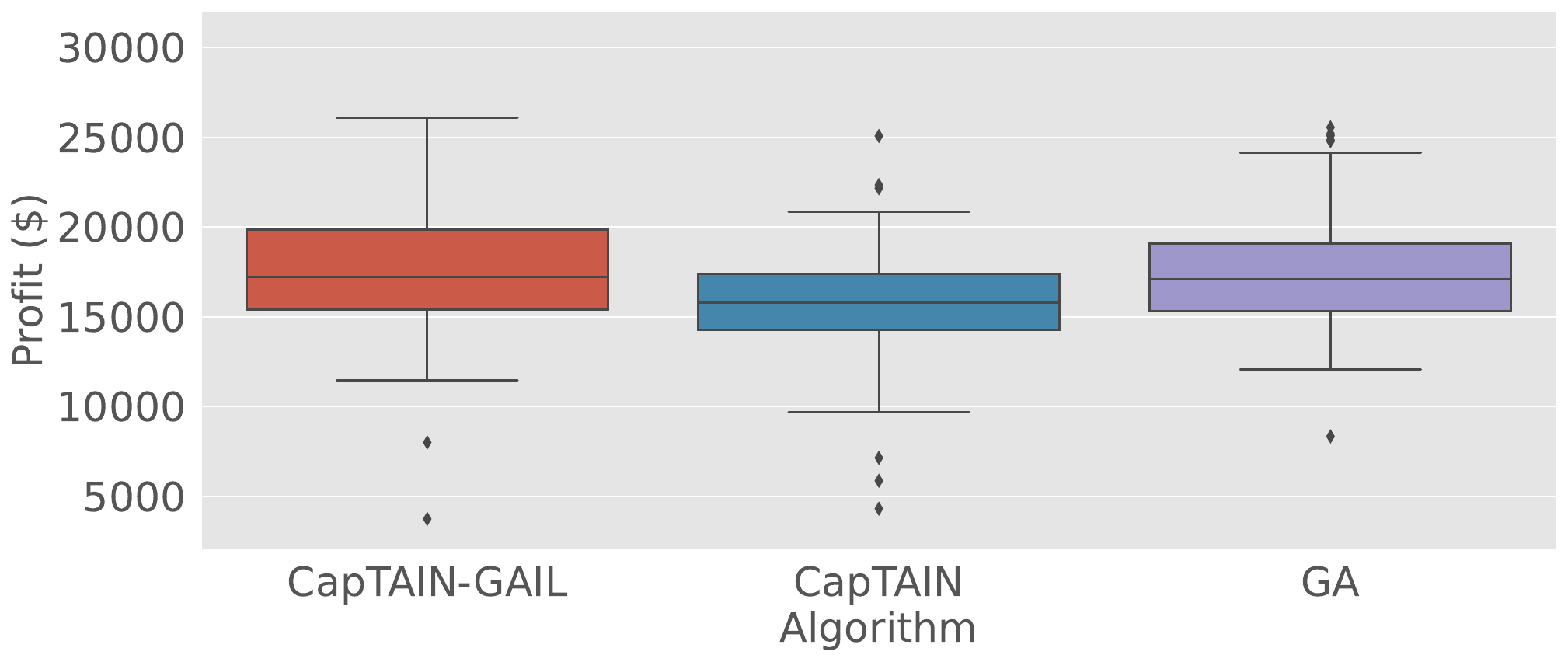}
    \caption{Comparing the average profits earned by all the 3 methods}
    \label{fig:compare_profit}
\end{figure}

Fig.\ref{fig:compare_profit} plots the average profits earned by all three methods, in the test scenarios that were a part of the expert demonstrations. It can be noticed that CapTAIN generates the least profit while CapTAIN-GAIL and GA almost earn equal profits on average. A more detailed analysis of the performance of the three methods is discussed below.

\subsubsection{CapTAIN-GAIL vs. CapTAIN}

\begin{figure}[hbt!]
    \centering
    \includegraphics[width=8.5cm]{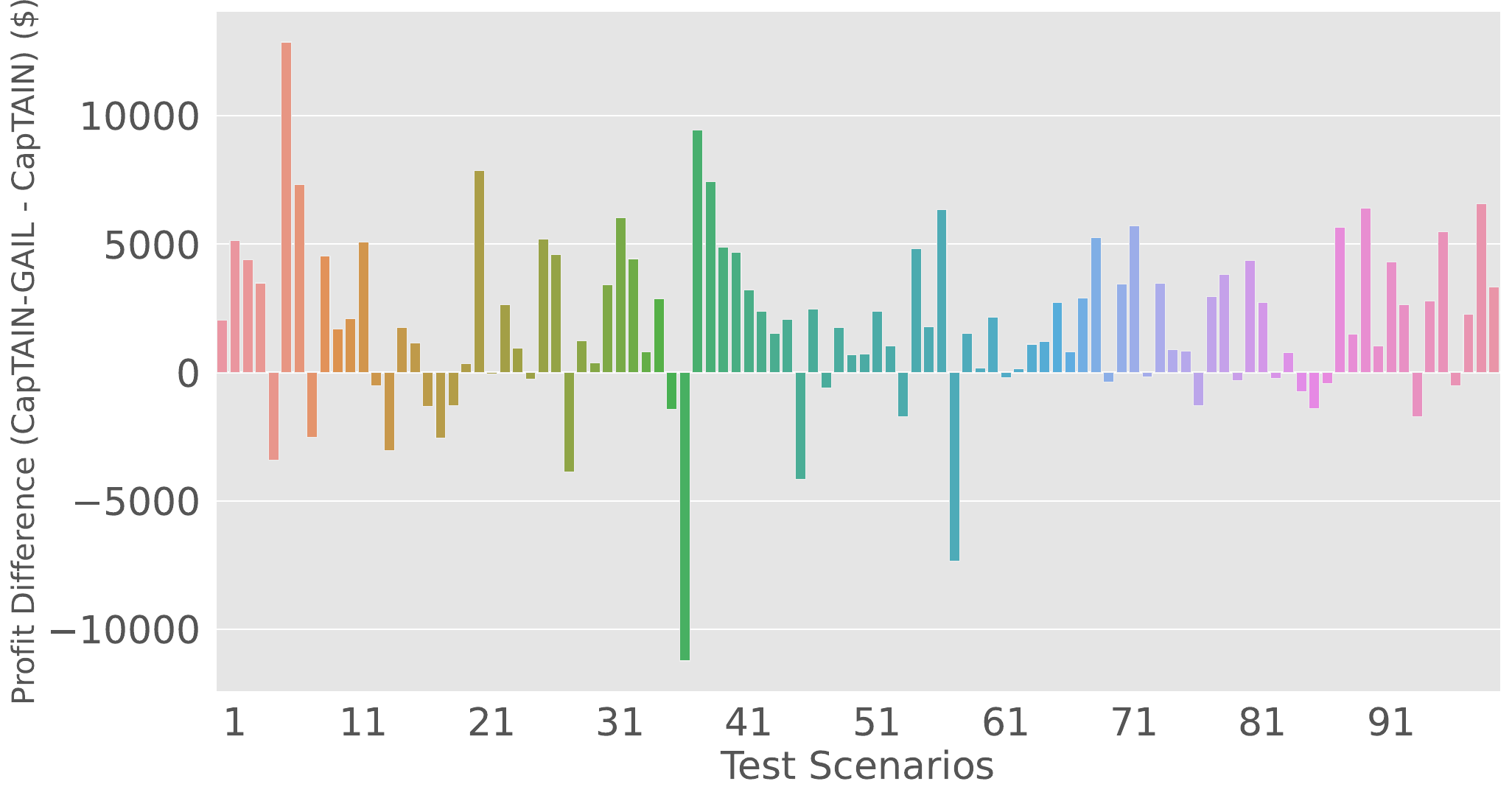}
    \caption{Difference in the profits earned by CapTAIN-GAIL and CapTAIN, Positive values indicate better performance by CapTAIN-GAIL and negative values indicate better performance by CapTAIN.}
    \label{fig:gail_vs_rl}
\end{figure}

Fig.\ref{fig:gail_vs_rl} compares the difference in the profits earned by CapTAIN-GAIL and CapTAIN. CapTAIN-GAIL generates more profits than CapTAIN in \textbf{73 out of 100 cases}, with the \textbf{average profit of CapTAIN-GAIL being $\bm{\$17587}$ while the average profit of CapTAIN is $\bm{\$15727}$}. To statistically evaluate the significance of this difference, we conduct a T-test with the null-hypothesis being that both the algorithms generate equal average profits. With a \textbf{p-value of $\bm{6.55\times 10^{-5}(<0.05)}$}, we have statistical evidence that CapTAIN-GAIL performs significantly better than CapTAIN.

\subsubsection{CapTAIN-GAIL vs. GA}

\begin{figure}[hbt!]
    \centering
    \includegraphics[width=8.5cm]{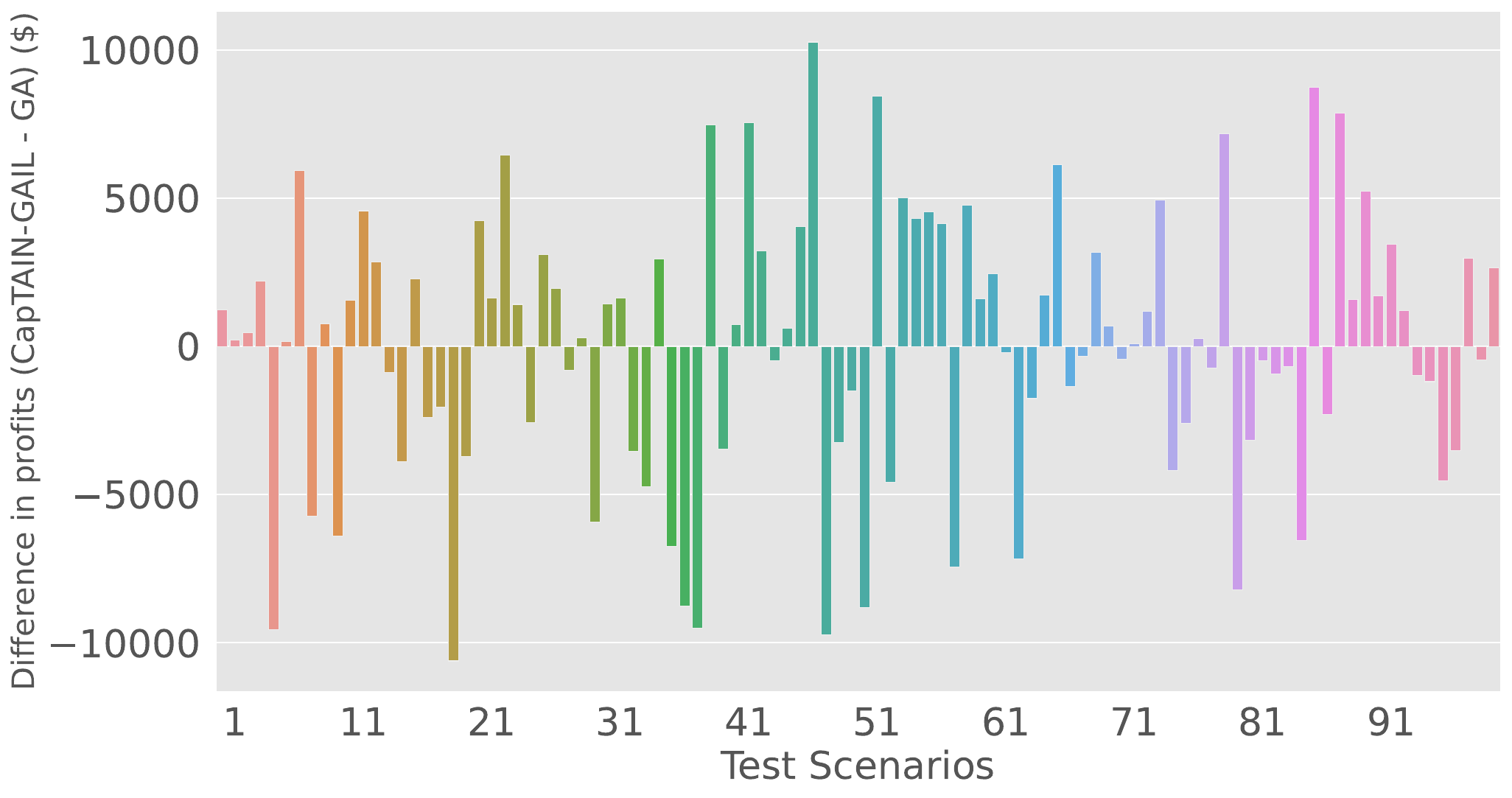}
    \caption{Difference in the profits earned by CapTAIN-GAIL and GA, Positive values indicate better performance by CapTAIN-GAIL and negative values indicate better performance by GA.}
    \label{fig:gail_vs_ga}
\end{figure}

Next, we evaluate the performance of CapTAIN-GAIL against GA. With Fig.\ref{fig:gail_vs_ga} plotting the difference in the profits earned by these two methods, we can see that CapTAIN-GAIL performs better than GA in \textbf{55 out of the 100 cases}, which is roughly half the number of test cases. The \textbf{average profit earned by GA is $\bm{\$17603}$ while the average profit earned by CapTAIN-GAIL is $\bm{\$17587}$}. Upon performing a T-test with the null hypothesis being that both methods earn the same average profit, we get a \textbf{p-value of $\bm{0.97(>0.05)}$} which indicates that statistically, both methods have a similar performance. This result matches our expectations since the imitation learning policy should at least perform similar to the expert when tested on the expert demonstrations.

{Even though both CapTAIN-GAIL and GA perform similarly in terms of the average profits they generate, it shall be noticed in Fig.\ref{fig:gail_vs_ga} that the difference in the profit values are quite large. This indicates that in cases where GA performs better, it outperforms CapTAIN-GAIL by a large margin while the opposite is also true, i.e. CapTAIN-GAIL outperforms GA by a large margin, in the cases where it performs better. One would ideally assume the difference in the profits to be close to 0, given that both the methods generate similar amount for profits, and this discrepancy in the performance of these two algorithms shall remain the subject of future research.}

\begin{figure}[hbt!]
    \centering
    \includegraphics[width=8.5cm]{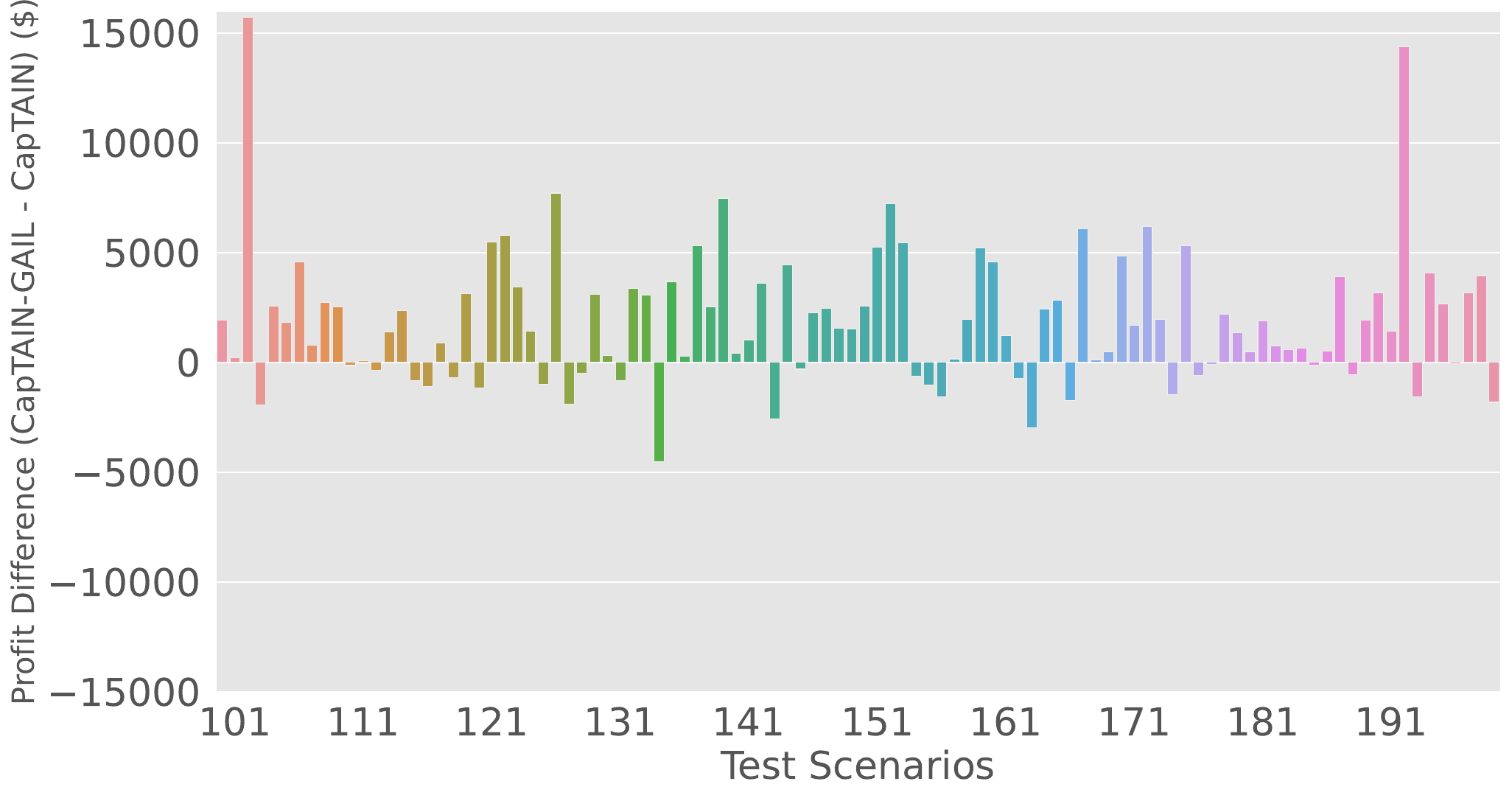}
    \caption{Difference in the profits earned by CapTAIN-GAIL and CapTAIN in completely unseen scenarios. Positive values indicate better performance by CapTAIN-GAIL and negative values indicate better performance by GA.}
    \label{fig:gail_vs_rl_unseen}
\end{figure}

\subsubsection{Generalizability Analysis}
To test the generalizability of CapTAIN-GAIL, we test it on a new set of 100 unseen scenarios that were not a part of the expert demonstrations on which it was trained, and compare it's performance against CapTAIN. Fig.\ref{fig:gail_vs_rl_unseen} plots the difference in the profits earned by CapTAIN-GAIL and CapTAIN. When compared to Fig.\ref{fig:gail_vs_rl}, we notice that CapTAIN-GAIL shows improvement in terms of performing better in cases where CapTAIN generates more profits (which can be seen by the reduced negative values in Fig.\ref{fig:gail_vs_rl_unseen}).

\begin{figure}[hbt!]
    \centering
    \includegraphics[width=8.5cm]{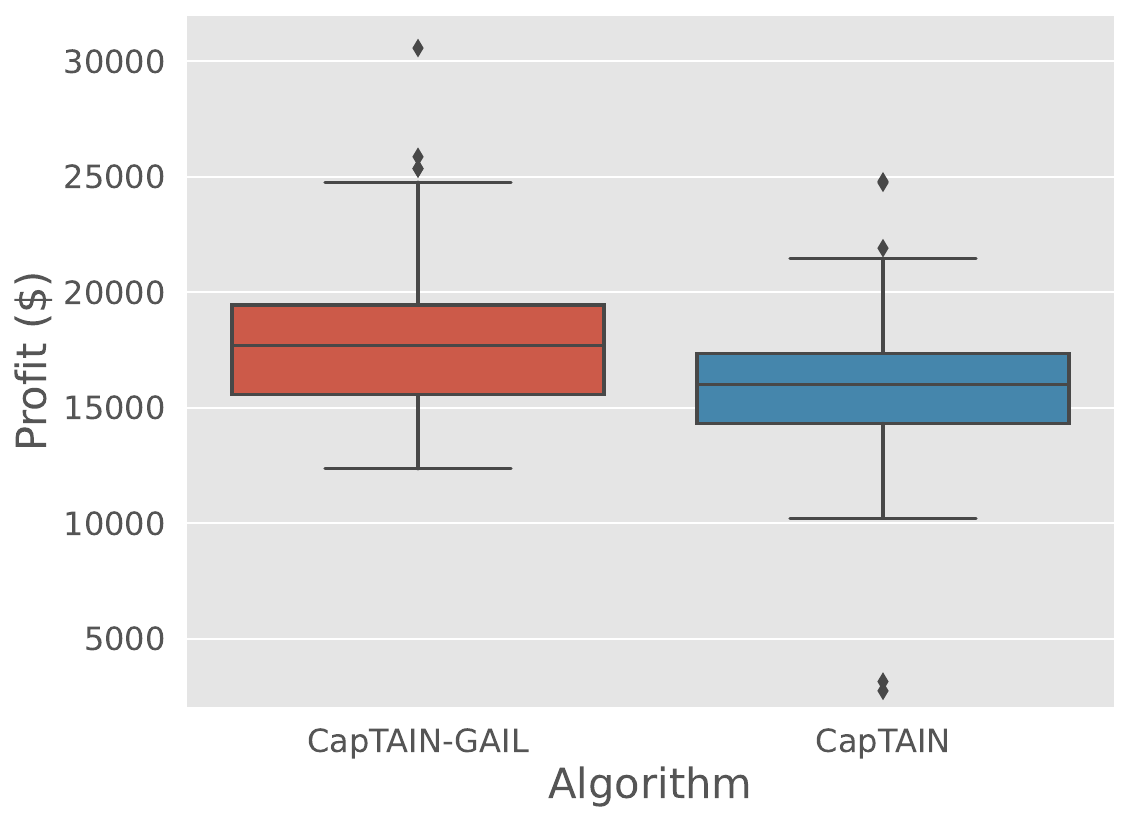}
    \caption{Comparing the average profits earned by CapTAIN-GAIL and CapTAIN in the unseen test cases.}
    \label{fig:compare_profit_unseen}
\end{figure}

Further analysing the mean profits earned by both the methods in the unseen scenarios (as shown in Fig.\ref{fig:compare_profit_unseen}) and comparing it to Fig.\ref{fig:compare_profit}, we notice that CapTAIN-GAIL achieves better mean performance and remarkably better bounding of the unseen worst-case scenarios (that can be seen in the reduce standard deviation between Fig.\ref{fig:compare_profit} and Fig.\ref{fig:compare_profit_unseen}), when compared to CapTAIN. In the unseen scenarios, \textbf{CapTAIN-GAIL generates an average profit of $\bm{\$17813}$} as compared to $\$17587$ in the expert demonstration cases while \textbf{CapTAIN generates $\bm{\$15902}$}. Additionally, the standard deviation of the profits earned, goes down from $\$3363$ in the expert demonstration cases to $\$3161$ in the unseen cases. This shows that CapTAIN-GAIL is generalizable across scenarios that are unseen in the expert demonstrations. Conducting a T-test with the null-hypothesis being that both the methods generate the same average profits, gives a \textbf{p-value of }$\bm{3.09\times 10^{-5}(<0.05)}$. Which means that CapTAIN-GAIL performs statistically better than CapTAIN.

\subsubsection{Further Analysis}
To further analyse the performance difference among the three methods, we look at their computation times as well as track the average number of idle decisions and flight decisions taken by the three methods.

The computing time for GA is calculated by adding the total time required to generate the solutions for an entire episode. Similarly, the computation times for CapTAIN and CapTAIN-GAIL are calculated based on the total forward-propagation time for the entire episode. It was found that GA took an average of $783.48 \pm 267.54$ seconds per episode while CapTAIN and CapTAIN-GAIL took $2.57 \pm 2.24$ seconds and $1.86 \pm 2.25$ seconds respectively. Thus, there is a significant advantage of using a policy trained using imitation learning when compared to an optimization solver like GA, in terms of computational time required.


\begin{figure}[hbt!]
\centering
    \begin{subfigure}[b]{0.5\textwidth}            
            \includegraphics[width=\textwidth]{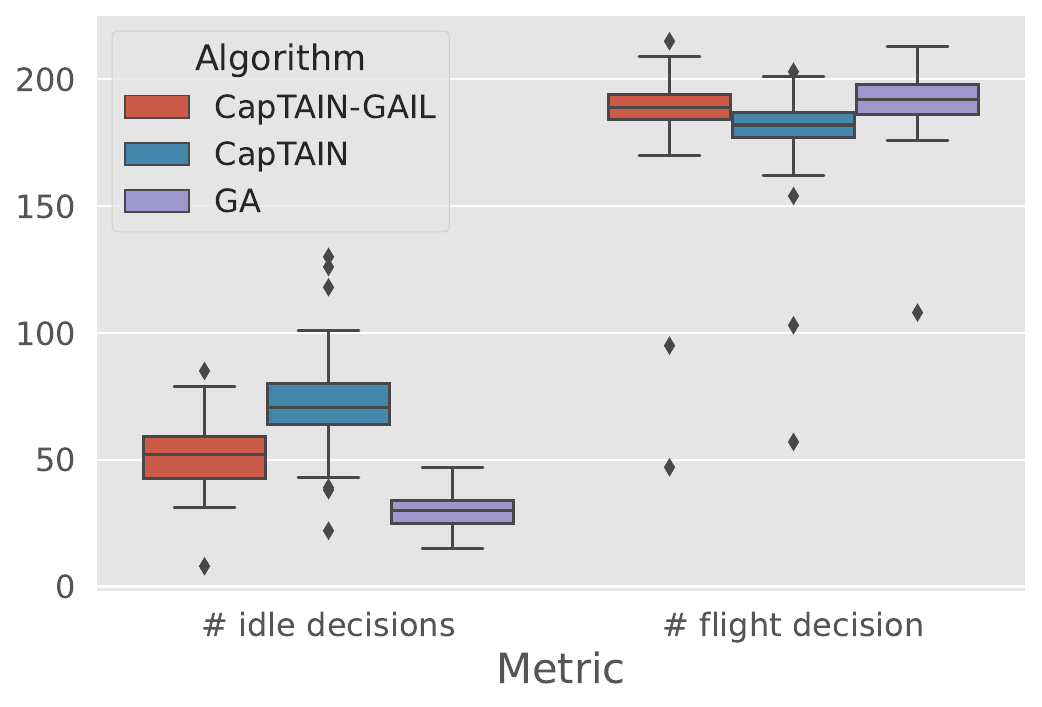}
            \caption{Comparing the number of idle decisions and number of flight decisions taken by the 3 algorithms.}
            \label{fig:flights}
    \end{subfigure}%
    \hspace{0.01cm}
    \begin{subfigure}[b]{0.5\textwidth}
            \centering
            \includegraphics[width=\textwidth]{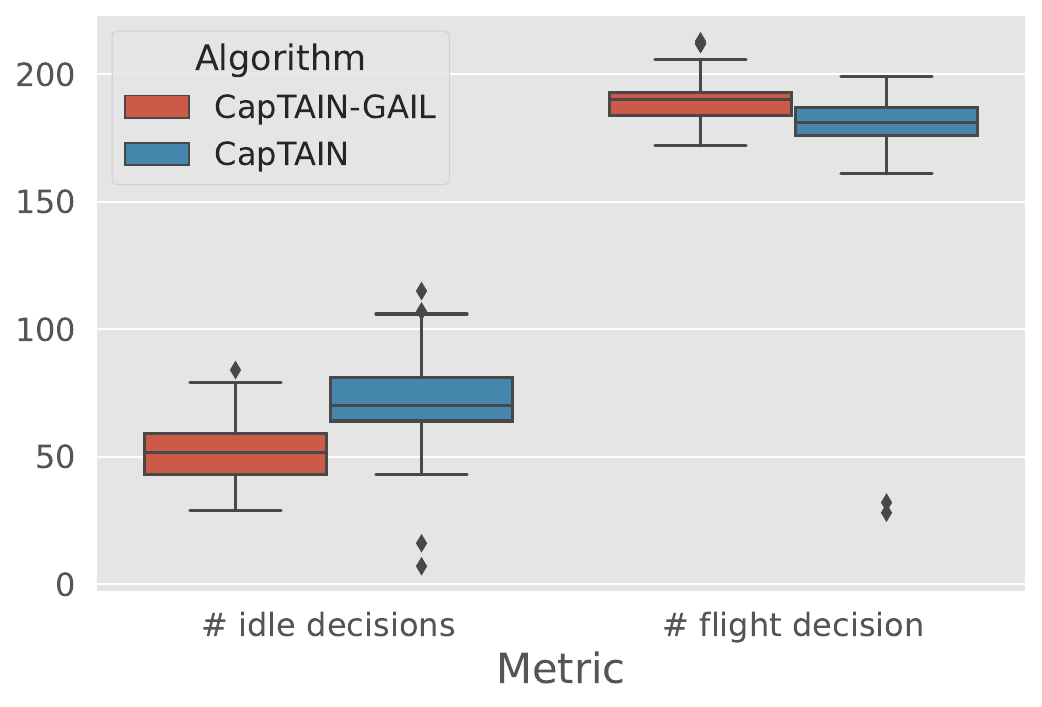}
            \caption{Comparing the number of idle decisions and flight decisions taken by the 2 algorithms in the unseen scenarios.}
            \label{fig:flights_unseen}
    \end{subfigure}
    \caption{Comparing the number of idle decisions and flight decisions taken by different algorithms under seen and unseen test scenarios.}
\end{figure}

Additionally, Fig.\ref{fig:flights} shows the average idle decisions and flight decisions taken by all the three algorithms in the scenarios present in the expert demonstrations. It can be noted that GA take the most number of flight decisions and the least number of idle decisions where as CapTAIN takes the least number of flight decisions and most number of idle decisions. Meanwhile, the decisions taken by CapTAIN-GAIL closely resemble those taken by GA, which is expected given that we have already seen that CapTAIN-GAIL performs similar to the GA.


In comparison, Fig.\ref{fig:flights_unseen} show the decisions taken by CapTAIN-GAIL and CapTAIN in the unseen test cases. It can be noted that the decisions remain fairly the same for both the algorithms. Additionally, the number of outliers in the unseen cases for CapTAIN-GAIL are lower than that for the expert demonstration cases. On an average, CapTAIN-GAIL takes 51.84 idle decisions and 187.29 fight decisions in the seen cases and 52.36 idle decisions and 189.48 flight decisions in the unseen cases, while CapTAIN takes 72.11 idle decisions and 179.96 fight decisions in the seen cases and 71.8 idle decisions and 178.39 flight decisions in the unseen cases. This further demonstrates the generalizability of CapTAIN-GAIL.

\section{Conclusion and Future Work}

In this paper, we developed CapTAIN-GAIL, a graph neural network based policy for Urban Air Mobility fleet scheduling, which presents itself as a challenging combinatorial optimization problem. CapTAIN-GAIL is trained using generative-adversarial imitation learning (GAIL), with (offline) expert demonstrations generated by a Genetic Algorithm or GA (that is impractical to be used directly/online due to its high computing cost). While prior work had shown that graph RL is uniquely capable of learning policies with performance clearly better than standard RL or other online approaches, there is still a significant optimality gap when compared with partly-converged results from a global optimizer such as GA. The imitation learning extension presented here (CapTAIN-GAIL) was found to successfully reduce this optimality gap, with the initial policy seeded by the earlier CapTAIN results. Our case studies involved scheduling the flights of 40 aircraft across 8 vertiports. When tested over 100 seen scenarios, CapTAIN-GAIL generated solutions with performance that is statistically similar to the solutions generated by the GA. Both across seen (during training) and unseen scenarios the new CapTAIN-GAIL results (in terms of the profit metric) are found to be significantly better than those by CapTAIN, thereby supporting our hypothesis of taking an imitation learning approach to reduce the optimality gap and improve performance. Notably CapTAIN-GAIL learnt to take more flight decisions than staying idle decisions for the UAM aircraft, compared to CapTAIN, which might be partly responsible for the improved performance. Interestingly, when comparing the performance of CapTAIN-GAIL against GA, for the small set of cases where their performance differed, the difference was quite high, the cause of which needs to be further explored in the future. Moreover, to allow more efficient use of costly expert demonstrations from GA, future work should look at approaches that can in situ identify when and where imitation is needed similar to the concept of adaptive sequential sampling in optimization, as opposed pre-computing a fixed set of expert demonstrations. 

\section*{Acknowledgments}
This work was supported by the Office of Naval Research (ONR) award N00014-21-1-2530 and the National Science Foundation (NSF) award CMMI 2048020. Any opinions, findings, conclusions, or recommendations expressed in this paper are those of the authors and do not necessarily reflect the views of the ONR or the NSF.
\vspace{-.3cm}





\bibliography{references}
\bibliographystyle{IEEEtran}

\end{document}